\documentclass{article}

\usepackage{arxiv}

\usepackage[utf8]{inputenc} 
\usepackage[T1]{fontenc}    
\usepackage{hyperref}       
\usepackage{url}            
\usepackage{booktabs}       
\usepackage{amsfonts}       
\usepackage{nicefrac}       
\usepackage{microtype}      
\usepackage{lipsum}
\usepackage{amsmath}
\usepackage{subcaption}
\usepackage{graphicx}
\usepackage[ruled,vlined]{algorithm2e}
\usepackage{bbm}
 \usepackage{url}
 
\renewcommand{\vec}[1]{\mathbf{#1}}
\newcommand{\matr}[1]{\mathbf{#1}}   

\DeclareMathOperator*{\argmin}{arg\,min}

\title{Formation of cell assemblies with iterative winners-take-all computation and excitation-inhibition balance}

\author{
    Viacheslav Osaulenko\\
    Igor Sikorsky Kyiv Polytechnic Institute \\
    \texttt{osaulenko.v.m@gmail.com} \\
\And
    Danylo Ulianych \\
    KyivAIGroup\\
    \texttt{d.ulianych@gmail.com} \\
}

\begin{document}
\maketitle

\begin{abstract}
This paper targets the problem of encoding information into binary cell assemblies. 
Spiking neural networks and k-winners-take-all models are two common approaches, 
but the first is hard to use for information processing
and the second is too simple and lacks important features of the first.
We present an intermediate model that shares the computational ease of kWTA and has more flexible and richer dynamics.
It uses explicit inhibitory neurons to balance and shape excitation through an iterative procedure. 
This leads to a recurrent interaction between inhibitory and excitatory
neurons that better adapts to the input distribution and performs 
such computations as habituation, decorrelation, and clustering. 
To show these, we investigate Hebbian-like learning rules
and propose a new learning rule for binary 
weights with multiple stabilization mechanisms.
Our source code is publicly available.
\end{abstract}


%
%
\section{Introduction}
Cell assembly is a group of repeatedly active interconnected neurons. 
It is one of the
 main candidates to represent and compute information in the brain \cite{buzsaki2010neural, palm2014cell} and a proper choice for creation of association memories \cite{Knoblauch2010}, pattern completion \cite{papadimitriou2020brain} and pattern recognition \cite{ahmad2019can}, 
 prediction of temporal sequences \cite{hawkins2016neurons},  
 and potentially representing casual probabilistic relationships \cite{barlow2001redundancy, rinkus2017radically}. 

Spiking neural networks and binary neural networks with discrete time are two broad classes of models that study cell assembly formation and transformation.
 A spiking neural network has rich and complex spatio-temporal dynamics, 
 but it is hard to show how neural activation patterns and a myriad of synaptic plasticity rules form the basis for information processing.
 Binary neural networks encode information into binary activations with a simple threshold model in discrete time.
They are much easier to analyze theoretically and computationally \cite{ahmad2016neurons}. 
 Particularly, one of the most common approaches is to use
 a k-winners-take-all (kWTA) model. kWTA is used for association memory models \cite{Graham1995ImprovingRF}, similarity search \cite{dasgupta2017neural}, fast and  efficient pattern recognition \cite{numenta2021}, and recently in the computation with assemblies framework \cite{papadimitriou2020brain}.
kWTA abstracts away inhibitory neurons and represents information as a vector with $k$ ones
(most active cells).
However, this model encodes input vectors into a fixed number of active neurons
, which contradicts biological neural networks (principles), where it is observed that the number of active neurons varies through time.
Moreover, a significant amount of evidence shows that inhibitory neurons not only select the top winners,
but also shape excitatory population activity \cite{isaacson2011inhibition}.

Here we present a model of cell assemblies formation that is on a higher abstraction level than spiking neural networks but computationally richer than simple threshold models. It explicitly models inhibitory neurons that balance excitation and provide additional mechanisms to control the activity of encoding cells.
The stabilization of two interacting populations requires an iterative procedure that we call iterative winners-take-all (iWTA). At each iteration, new active neurons are formed. This process resembles latency coding and abstracts away precise spike timings. Cell assemblies are formed from all active cells during the restoration of the excitation-inhibition balance (see fig.\ref{fig:iterations}).

Interestingly, we receive a paradoxical effect reported previously \cite{Sadeh2020}, where an increase in input excitation might decrease the population activity. 
Similar to kWTA, the iWTA model preserves similarity (encodes similar inputs into similar outputs).
But in addition, it is able to change the number of active cells according to the input distribution like in habituation. Also, it can selectively change the overlap between different cell assemblies (e.g. decorrelation) to better reflect the bottom-up or top-down signals. 
Furthermore, it reduces noise in the clusters of noisy data.

To show this, we present Hebbian-like learning rules for binary $\{0,1\}$ weights and activations, supplemented with stabilization and pruning mechanisms. More complex and interesting dynamics require more parameters, making the iWTA model computationally expensive.  
Still, the investment is worth making, as we show in our results.

%
%
\section{A common k-winners-take-all (kWTA) model}
It is widely recognized that a simple threshold model of a neuron does not reflect 
richness and complexity of biological neuron computation 
\cite{Poirazi2003PyramidalNA, Smith2013DendriticSE}. 
Dendritic spatio-temporal processing requires more complex approaches
\cite{London2005DendriticC, Jones2021MightAS}. 
Still, a threshold model and its variations remain influential. 
The vectorized form of a threshold model can be written as:

\begin{equation}
\vec{y} = \theta(\matr{w}\vec{x} - \vec{t})
\end{equation}

The input vector $\vec{x}$  (bold notation) is multiplied 
by a matrix of weights $\matr{w}$ and the result is compared with a vector of thresholds $\vec{t}$.
Theta function is defined as $\theta(x) = \{1, \text{ if } x \geq 0; \thinspace 0, \text{ otherwise}\}$
and applied element-wise to vector components. 
This model defines activation of the neuronal population $\vec{y}$,
and $t_i$ can be different for each neuron $i$.

Multiple Hebbian-like rules exist to learn weights,
however Hebbian theory tells little or none about the threshold.  
One way to treat the threshold is to consider it as a parameter that optimizes some loss function, like in classic artificial neural networks. Another, more biologically plausible way, is to define and adjust the threshold via interactions with inhibitory neurons. 
The simplest and most widely used form of such interaction is a k-winners-take-all model.
It implicitly models inhibitory neurons by dynamically adjusting the threshold to select exactly $k$ active neurons:

\begin{equation}
\label{eq:kWTA}
\begin{aligned}[c]
\vec{y} &= \theta(\matr{w}\vec{x} - \vec{1}t) \\
t & = \argmin_t{\left|\sum_i{y_i(t)} - k \right|}
\end{aligned}
\qquad\Longleftrightarrow\qquad
\begin{aligned}[c]
\vec{y} &= \text{kWTA}(\matr{w} \vec{x}, k)\\
\end{aligned}
\end{equation}

Here the threshold $t$ is the same for all neurons ($\vec{1}$ is a vector of ones) and is selected such that to make vector $\vec{y}$ with exactly $k$ ones (hence k-winners-take-all, see fig.\ref{fig:network_scheme}).
Usually, this is done by sorting the array and selecting the top $k$ largest elements.  
For example, $\text{kWTA}([1,2,3,4], 2) = [0,0,1,1]$ results in a binary vector with the sparsity equals to 0.5. We define the sparsity of a vector $\vec{y}$ as $s_y = \frac{a_y}{N_y}$, where $a_y=||y||_0$ is the number of ones and $N_y$ is the vector size.

kWTA is simple and fast to compute \cite{Maass2000OnTC}.
It is not necessary to sort the array to select the top $k$ largest values. Sorting can be substituted by a recurrent neural network that implements this function  \cite{yang1997dynamic, tymoshchuk2009discrete, mao2007dynamics}.
kWTA models processing in simple biological systems like computation in Kenyon cells of a fruit fly olfactory system that receives inhibition from one cell only \cite{dasgupta2017neural}.
Random weights initialization is a special and very interesting case. Following the Johnson–Lindenstrauss lemma, kWTA with random weights preserves similarity \cite{dasgupta2017neural} - similar inputs are encoded in similar outputs.
This can be interpreted as local sensitive hashing for fast similarity search \cite{rachkovskij2015formation}. Moreover, when $N_y \gg N_x$, kWTA preserves almost all information about the input, which can be easily restored as $\vec{x}=\text{kWTA}(\matr{w}^T\vec{y}, a_x)$  \cite{osaulenko2020binary}.

However, this simple activation function poses its drawbacks and limitations. 
It keeps the encoding sparsity fixed for all inputs. Therefore, it is not possible to achieve habituation, where the encoding sparsity reflects the frequency of the input, like in biological systems.
Also, kWTA cannot be applied in a predictive coding framework, 
when a predicted stimulus is encoded with fewer active neurons \cite{Spratling2017ARO}.
Furthermore, there is multiple evidence showing that inhibitory neurons not only stabilize the excitatory populations but also shape them \cite{isaacson2011inhibition, barron2017inhibitory}. Inhibition plays a direct role in neural computation and cannot be abstracted away.

%
%
\section{Iterative winners-take-all (iWTA) model}
Multiple prior works showed the importance of explicit inhibition 
\cite{brunel2000dynamics, barron2017inhibitory} that refines the selectivity of excitatory neurons, for example in case of visual \cite{ozeki2009inhibitory}
and olfactory systems \cite{yu2014sparse}. Other works used only negative recurrent connections between excitatory neurons \cite{spratling1999pre}, and in some cases it forms sparse encoding \cite{foldiak1990forming}.
However, a separate inhibitory population gives finer control and introduces new behavior, for example, a \textit{paradoxical effect} \cite{tsodyks1997paradoxical, Sadeh2020},  where an increase in the input to the inhibitory neurons in a strongly coupled excitatory-inhibitory network might decrease their population activity.
 However, most of these models use real-valued activations or complex dynamics modeled with differential equations that are hard to link to cell assemblies. 
There is a need for a model that shares the computational ease of kWTA and has more flexible and richer dynamics. Next, we present such a model.

\begin{figure}[t!]
  \begin{subfigure}[t]{0.58\textwidth}
        \includegraphics[width=\linewidth]{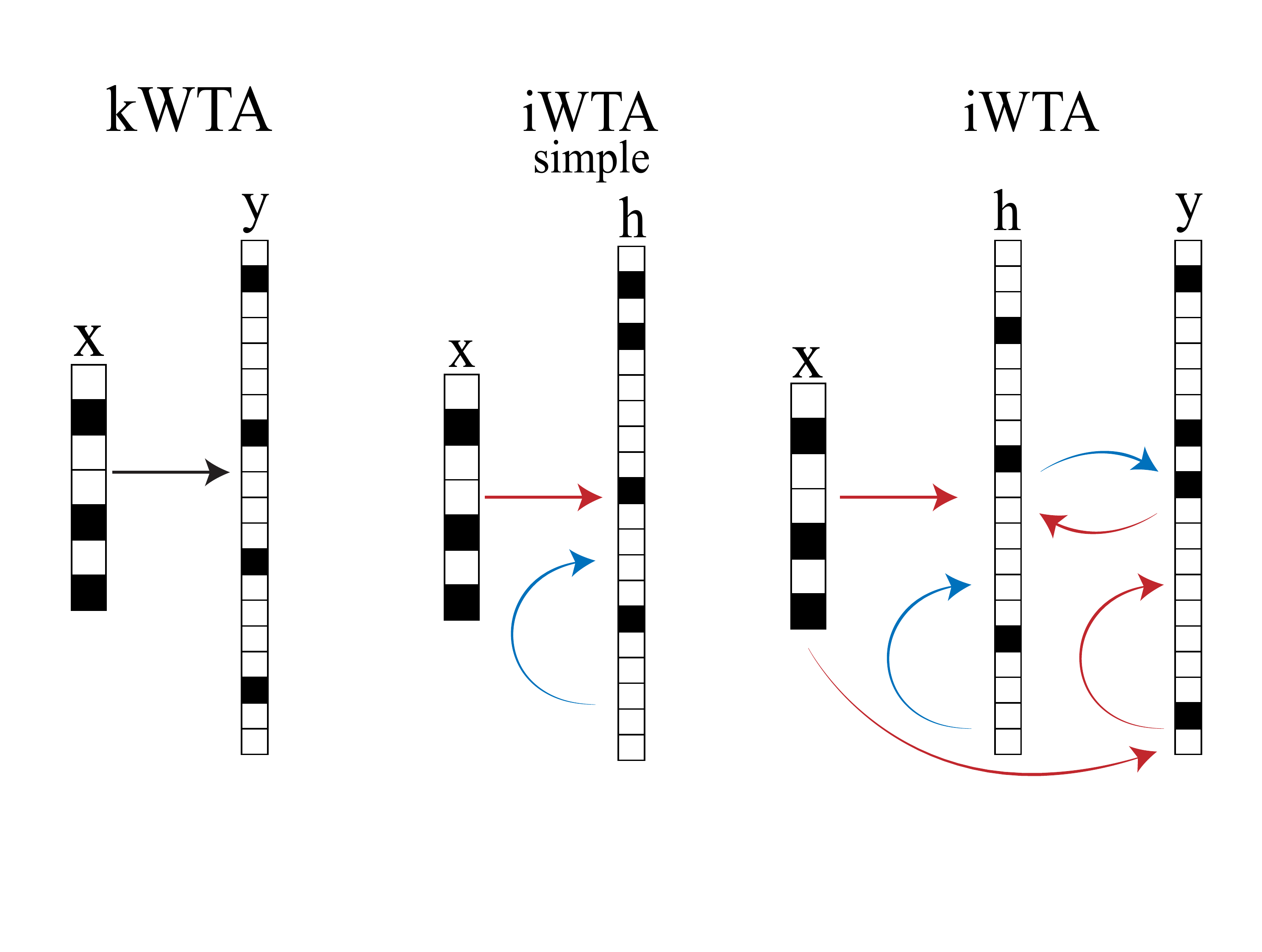}
      \caption{\label{fig:network_scheme} }  
  \end{subfigure}
   \hfill
  \begin{subfigure}[t]{0.4\textwidth}
        \includegraphics[width=\linewidth]{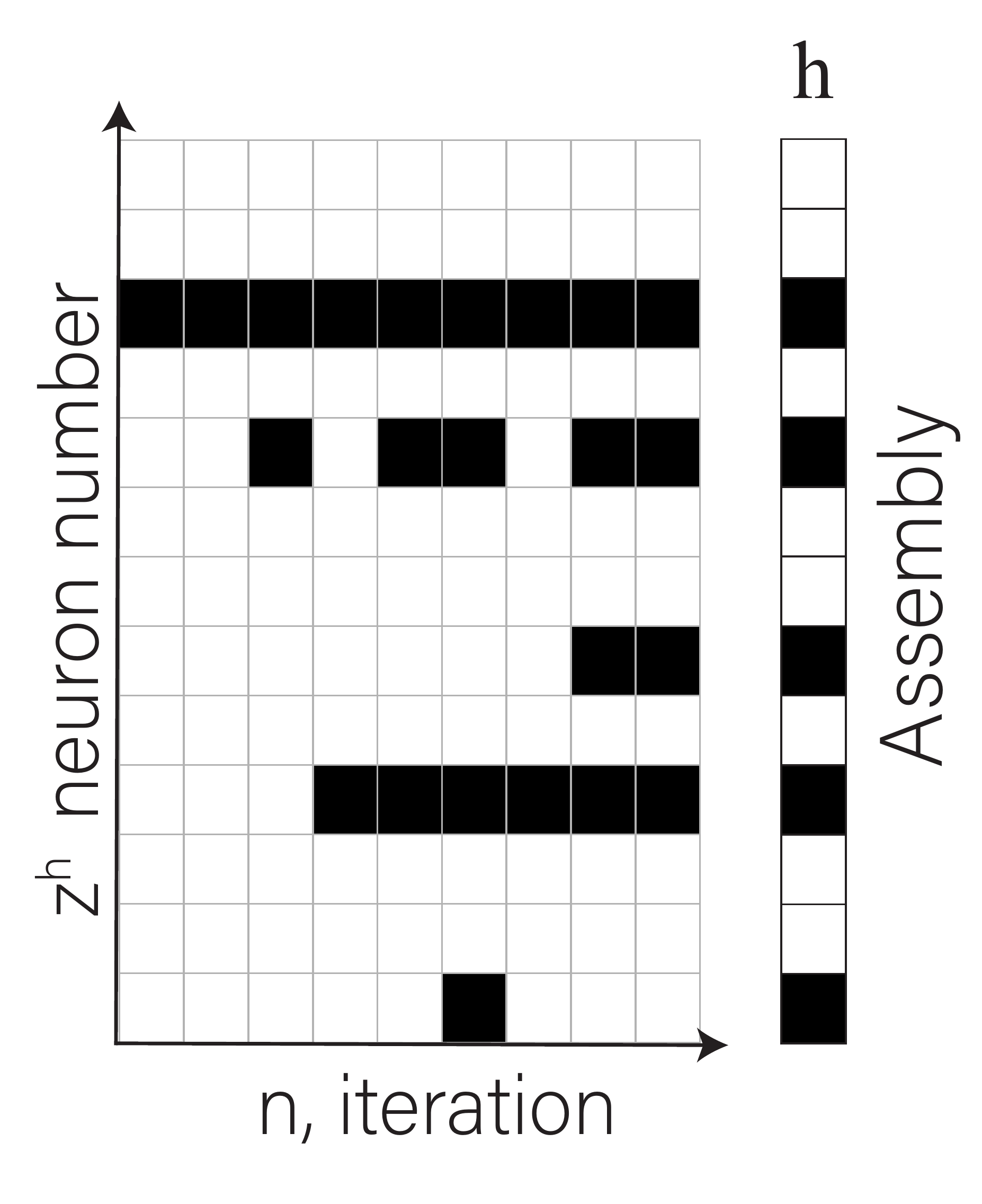}
      \caption{\label{fig:iterations}}
  \end{subfigure}  
         \caption{
     	a) An illustration of different architectures for encoding an input vector into a binary vector. Left: a kWTA model, a single parameter $k$ controls the sparsity of the encoding. Middle: a simple iWTA model, self-inhibitory connections control the sparsity of the encoding. Right: the iWTA model, the input vector is encoded into excitatory $\vec{y}$ and inhibitory $\vec{h}$ populations through iterative procedure. Excitatory connections are shown in red, inhibitory - in blue.
     	b) Active inhibitory neurons (black) at different iterations. With each iteration, more and more neurons become active. Self-inhibition may deactivate some neurons from previous iterations. The resulting encoding is assembled from active neurons from all iterations.
                 	}
\end{figure}

Consider at first a simplified version where the input $\vec{x}$ excites population $\vec{h}$ that forms self-inhibitory connections (see fig.\ref{fig:iterations} middle). 
Cells with the largest excitation are activated first by setting the threshold to the maximal value.
At the next iteration, the threshold is decreased, and new cells with lower excitation become active.
However, this time, neurons receive inhibition from active neurons from previous iterations. 
The iterative procedure continues until the threshold reaches zero, that is when the inhibition balances excitation. 
The resulting encoding vector $\vec{h}$ combines all active cells across the iterations. 

More formally, a simplified iterative winners-take-all model encodes an arbitrary input binary vector $\vec{x} \in \{0,1\}^{N_x}$ of size $N_x$ into a binary vector $\vec{h} \in \{0,1\}^{N_h}$ using random matrices $\matr{w^{xh}} \in \{0,1\}^{N_h \times N_x}, \matr{w^{hh}} \in \{0,1\}^{N_h \times N_h}$ according to:

\begin{equation}
\label{eq:iwta_simple}
\begin{aligned}[c]
\vec{z^h_n} & = \theta(\matr{w^{xh}}\vec{x} - \matr{w^{hh}}\vec{h_n}  - \vec{1}t_n) \\ 
\vec{h_{n+1}} & = \bigcup\limits_{i=1}^{n} \vec{z}^h_{i} \\
t_{n+1} & =  t_{n} - 1 \\
\end{aligned}
\qquad\Longleftrightarrow\qquad
\begin{aligned}[c]
\vec{h} &= \text{iWTA}(\matr{w^{xh}}\vec{x} - \matr{w^{hh}}\vec{h})\\
\end{aligned}
\end{equation}

The initial values are set to $t_0 = \max(\matr{w^{hx}}\vec{x})$ and $\vec{h_0} = \vec{0}$. 
The weight matrices are initialized randomly such that each row contains a fixed number of ones (sampling from a Bernoulli distribution is another choice). We use the notation $\matr{w}^{pq}$ to denote the weights from a layer $p$ to a layer $q$.  

$\vec{z^h_n}$ represents active neurons at the $n$-th iteration that are determined by the input excitation $\matr{w^{xh}}\vec{x}$ and self-inhibition $\matr{w^{hh}}\vec{h}$. Fig.\ref{fig:iterations} shows the neuronal activation $\vec{z^h_n}$ at different iterations.
The neurons in $\vec{z^h_n}$ can fire always, occasionally or only once.
The resulting encoding accumulates intermediate $\vec{z^h_n}$ vectors into $\vec{h}$  via logical OR operation.

Note, our model is similar to latency coding where neurons that receive larger excitation emit spikes first.  Similarly, in fig.\ref{fig:iterations} neurons with larger input become active faster and resemble a spike raster plot. The iterative procedure can be considered as an abstraction of spiking dynamics that forms cell assemblies from active cells while the excitation-inhibition is balancing.

The sparsity ($s_h$) of the encoding vector $\vec{h}$ is determined by the matrices
$\matr{w^{hh}},\matr{w^{hx}}$ and the vector $\vec{x}$. The more ones in $\matr{w^{hh}}$, the stronger the inhibition, the faster it balances excitation and the fewer active neurons are in $\vec{h}$ ($s_h$ becomes smaller). Note, self-inhibition is crucial for the model to converge.
Importantly, a simple iWTA model compared to a kWTA model uses an additional inhibitory matrix with $N_h^2$ parameters. 
This requires more memory and computation but gives an advantage. The inner product $\matr{w^{hh}}\vec{h}$ can be treated as additional threshold $t_h$ that depends implicitly on the input vector $\vec{x}$.
Therefore, learning the inhibition matrix adapts the sparsity of encoding to individual inputs that mimic an increase or decrease in the threshold.
On the contrary, the encoding sparsity of a kWTA model is kept fixed. A similar idea to use an input-dependent threshold to improve kWTA was proposed recently in \cite{dasgupta2020expressivity}. 

The case with only self-inhibitory connections is simpler and easier to understand. But association memory and linking cell assemblies through time require excitatory connections. Therefore, a full iWTA model encodes input to both excitatory and inhibitory neural populations.  Fig.\ref{fig:network_scheme} on the right shows a schematic network architecture; excitatory and inhibitory connections are shown in red and blue respectively.  

Iterative procedure starts by initializing the excitatory $\vec{y_0}$ and inhibitory $\vec{h_0}$ populations with zeros. The threshold is set to $t_0 = \max(\matr{w^{xh}}\vec{x}, \matr{w^{xy}}\vec{x})$ and the algorithm proceeds as follows:

\begin{equation}
\label{eq:iwta}
\begin{aligned}[c]
\vec{z^y_n} & =  \theta(\matr{w^{xy}}\vec{x} - \matr{w^{hy}}\vec{h_n} + \matr{w^{yy}}\vec{y_n}  - \vec{1}t_n) \\
\vec{z^h_n} & =  \theta(\matr{w^{xh}}\vec{x} - \matr{w^{hh}}\vec{h_n} + \matr{w^{yh}}\vec{y_n}  - \vec{1}t_n) \\
\vec{y_{n+1}} & = \bigcup\limits_{i=1}^{n} \vec{z^y_i} \hspace{30pt} \vec{h_{n+1}} = \bigcup\limits_{i=1}^{n} \vec{z^h_i} \\ 
t_{n+1} & =  t_{n} - 1 \\
\end{aligned}
\qquad\Longleftrightarrow\qquad
\begin{aligned}[c]
\vec{y} &= \text{iWTA}(\matr{w^{xy}}\vec{x} - \matr{w^{hy}}\vec{h} + \matr{w^{yy}}\vec{y})\\
\vec{h} &= \text{iWTA}(\matr{w^{xh}}\vec{x} - \matr{w^{hh}}\vec{h} + \matr{w^{yh}}\vec{y})\\
\end{aligned}
\end{equation}

The iterative procedure converges to stable binary vectors $\vec{y}, \vec{h}$. By changing the input or weight matrices, the model finds a new stable configuration. 
The model requires six weight matrices but is similar to a simple iWTA eq.\ref{eq:iwta_simple}. With a recurrent excitation $\vec{y} \rightarrow \vec{y}$, the model can implement association memory (pattern completion or association with another pattern). Also, more parameters mean higher model expressivity, and the encoding can better reflect the input distribution. 

By learning different subsets of the weights, we can adapt the encoding to particular inputs, changing the encoding sparsity and intersection with other encodings (see section \ref{learning}). Before that, we analyze how the encodings are influenced by the input and weight sparsity in cases when the connections are random and fixed.


\begin{figure}[t!]
  \label{fig:weight_sparsity}
  \begin{subfigure}[t]{0.48\textwidth}
        \includegraphics[width=\linewidth]{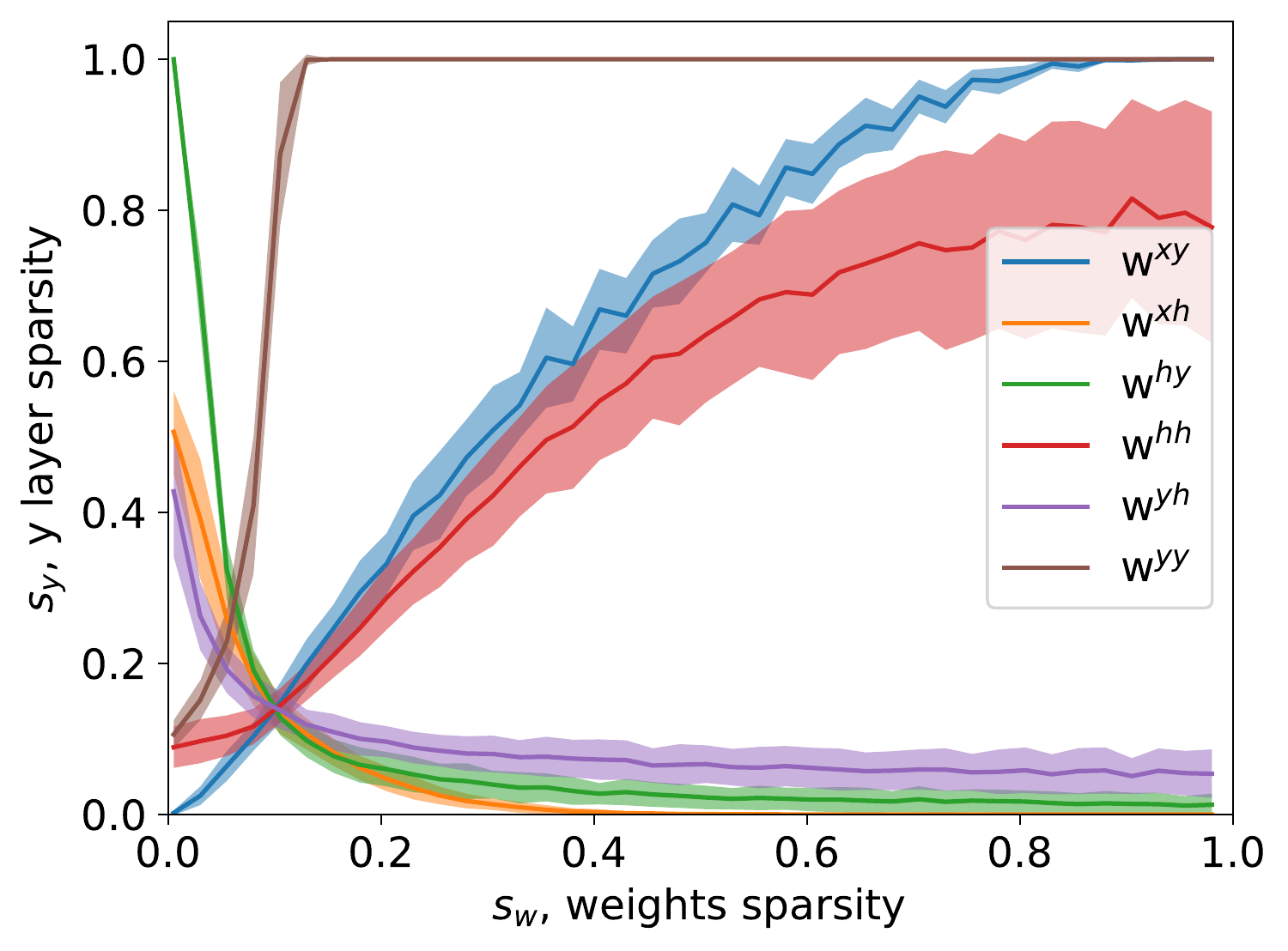}
      \caption{\label{fig:s_y_s_w} }  
  \end{subfigure}
   \hfill
  \begin{subfigure}[t]{0.48\textwidth}
        \includegraphics[width=\linewidth]{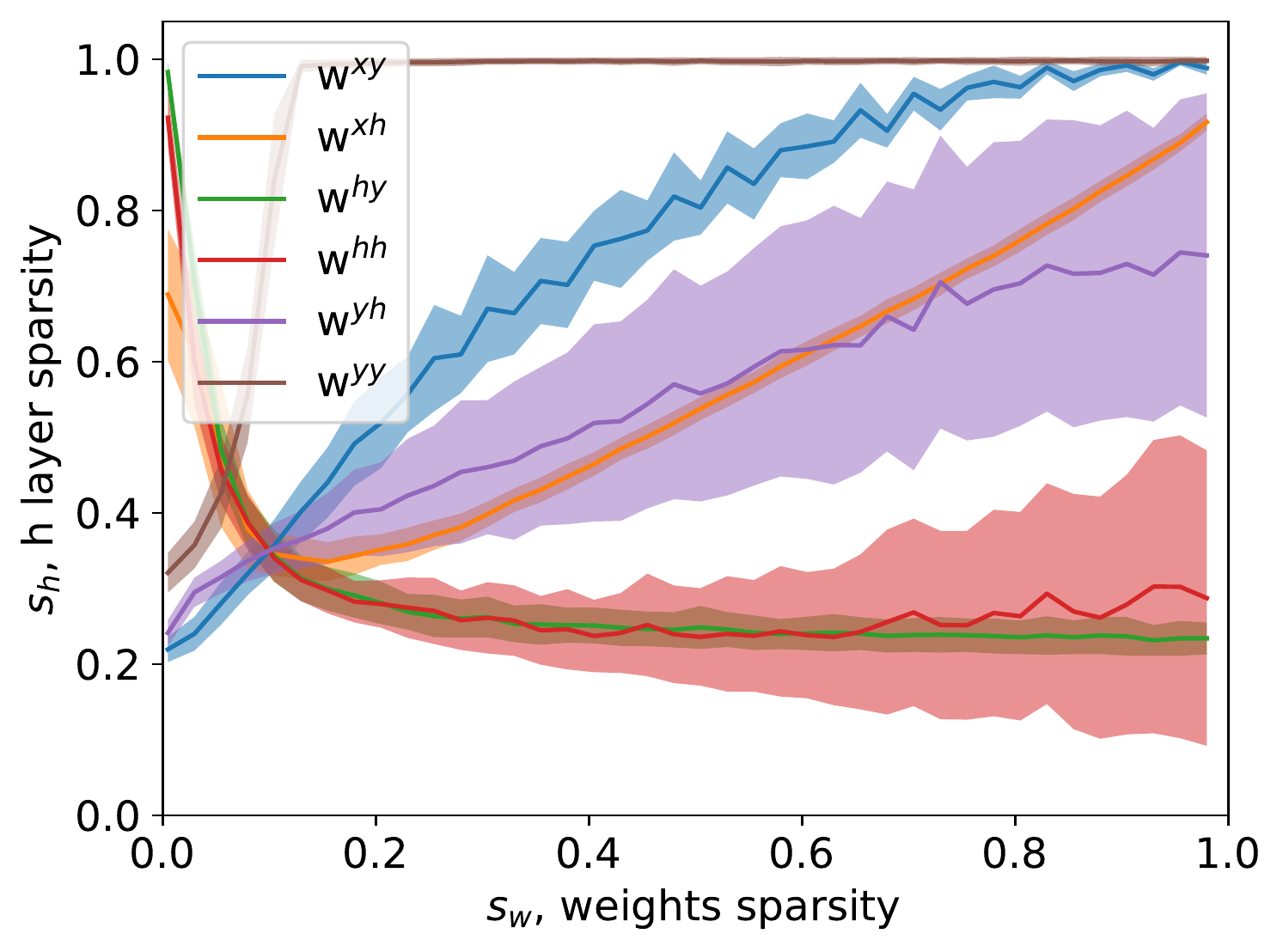}
      \caption{\label{fig:s_h_s_w}}
  \end{subfigure}  
         \caption{
             	Dependence of the sparsity of encodings on the weight sparsity. 
     	a) The number of active excitatory cells rises with an increase in the number of connected synapses for three different weights - $\matr{w^{yy}}, \matr{w^{hh}}, \matr{w^{xy}}$, - and lowers for $\matr{w^{xh}}, \matr{w^{hy}}, \matr{w^{yh}}$.
     	b) The sparsity of inhibitory cells behaves differently compared to excitatory cells. $\matr{w^{hy}}$, $\matr{w^{hh}}$ connections decrease and $\matr{w^{xy}}$, $\matr{w^{yh}}$, $\matr{w^{yy}}$  increase the sparsity of $\vec{h}$, and $\matr{w^{xh}}$ connections influence $s_h$ non-monotonically.
         	}
\end{figure}

\subsection{How the model controls sparsity}
We initialize weight matrices such that each row has a fixed number of randomly distributed ones (each neuron has the same number of connected synapses).
The sparsity of a matrix is defined as the number of non-zero weights divided by its size.

Fig.2 shows how the sparsity of six weight matrices influences the sparsity of excitatory (left plot) and inhibitory (right plot) populations. 
For each curve, we change the number of nonzero weights per row ($a_{pq}$) of one matrix keeping all others fixed.
The default network parameters are $a_{xy}= 20, a_{xh}= 20,a_{hy}= 20,a_{hh}= 20,a_{yh}= 20,a_{yy}= 5, N_x=200, N_y=200, N_h=200$. The input vector $\vec{x}$ is generated randomly with $a_x=20$ ones and the output sparsity is averaged over different trials. Both figures show a standard deviation of sparsity values as shaded areas around each curve. The same experiment with random weights sampled from a Bernouilli distribution shows similar results but with higher variance.

Interaction of excitatory and inhibitory populations shows peculiar dynamics: the sparsity of the encoding layer $\vec{y}$ rises with the direct increase in the connectivity of $\matr{w^{xy}}, \matr{w^{yy}}$ but also via the decrease in an inhibitory population by growing the self-inhibition ($\matr{w^{hh}}$).
Similarly, for layer $\vec{h}$, an increase in connectivity from the input and the excitatory layer $\vec{y}$ increases $s_h$. However, as the input connections increase $\matr{w^{xh}}$ (orange curve in fig.\ref{fig:s_h_s_w}), at first we observe a decrease in the activity of the inhibitory layer.  But later the input excitation drives more inhibitory neurons to be active.
    This paradoxical effect was also observed in \cite{tsodyks1997paradoxical, sadeh2021excitatory} and results from nonlinear excitatory-inhibitory interaction.

Also, we performed an experiment where the input sparsity changes while the connections are random and fixed. The sparsity of excitatory and inhibitory populations depends nonlinearly on the input sparsity. As the number of active input neurons increases, the sparsity of $\vec{h}$ and $\vec{y}$ decreases. At some point, with a higher $s_x$, the sparsity of $\vec{h}$ and $\vec{y}$ starts to increase and saturates to 1. Interestingly, for some values of weights sparsity, we observed the case when $s_y$ remained constant on a wide range of input sparsity, while $s_h$ is increased to compensate for larger input excitation.
    Different layer sizes ($N_x, N_y, N_h$) almost do not influence $s_y(s_x)$ and $s_h(s_x)$,
    in case when the number of ones in the weights is fixed. See fig.1 in supplementary material \cite{Github2021}.

Another experiment showed that the iWTA model preserves input similarity like kWTA.
We generated different pairs of input vectors ($\vec{x_1},\vec{x_2}$) with equal sparsity and applied the iWTA model.
The output cosine similarity  $\cos(\vec{y_1},\vec{y_2})$ depends almost linearly on the input similarity ($\cos(\vec{x_1},\vec{x_2})$) for high $s_y$. The lower $s_y$, the more nonlinear the dependence, the less similarity is preserved. The kWTA model has the same dependence. See fig.2 in the supplementary material \cite{Github2021}. 

Learning rules change weight sparsity differently for different neurons.
For example, the connections from $\vec{h}$ to $\vec{y}$ inhibit frequent stimuli stronger than rare stimuli, resulting in fewer active output neurons, 
similarly to habituation.
Alternatively, this can be achieved with an increase in $s_{xh}$ or $s_{yh}$ for more frequent inputs (see fig.\ref{fig:s_y_s_w}).
In addition, other effects like decorrelation and clustering can be achieved, however, learning rules should be supplemented with multiple stabilization mechanisms. See details in the next section.

%

\section{Learning rules for cell assemblies with binary weights}
\label{learning}
Hebbian-like learning rules have undergone numerous modifications since their establishment in neuroscience \cite{citri2008synaptic}.  
Their form depends on the coding strategy, either it is a family of STDP rules for spike coding \cite{markram2012spike, vogels2011inhibitory} or correlation-based rules for rate coding \cite{dayan2001theoretical}, or threshold models for sparse binary codes \cite{cui2017htm}.
A particularly interesting case is when the input, the output, and the weight matrix are binary (filled only with zeros and ones). The common approaches to deal with binary weights include Bayesian probabilistic methods like Monte Carlo sampling \cite{huang2014neurons}, gradient descent methods \cite{rastegari2016xnor, courbariaux2016binarized}, and mentioned threshold models. 
In our work, we follow the latter approach and describe several local Hebbian 
learning rules applicable for the iWTA eq.\ref{eq:iwta}. 
The source code for these learning rules applied to our iWTA model is publicly available at \cite{Github2021}.

\subsection{Simple Hebbian rule without forgetting}

The simplest way to learn connections between two binary vectors $\vec{x_\text{pre}}$ and $\vec{x_\text{post}}$ in a Hebbian-like manner via a binary matrix $\matr{w}$ is, perhaps, the Willshaw associative memory model \cite{willshaw1969non}:  $\matr{w^{ij}} = 1, \text{ if } \vec{x_\text{post}}^i = 1 \text{ and } \vec{x_\text{pre}}^j = 1$. We use a similar approach written in the vector form:

  \begin{equation}
  \label{eq:willshaw}
  \begin{aligned}
        \matr{w} = \theta(\matr{w} + (\vec{x_\text{post}} \otimes \vec{x_\text{pre}}) \cdot \matr{m})
  \end{aligned}
  \end{equation}

 where $\otimes$ denotes the outer product and $\matr{m}$ is a boolean mask that chooses which connections from the outer product to update to avoid quick saturation of a learned matrix. This mask determines the speed of a learning process. The vectors $x_\text{pre}$ and $x_\text{post}$ denote activations of pre- and postsynaptic layers of neurons respectively and can be one of $\vec{x}$, $\vec{h}$, or $\vec{y}$. For example, for the inhibitory weights $\matr{w^{hy}}$, eq.\ref{eq:willshaw} takes the following form: $\matr{w^{hy}} = \theta(\matr{w} + (\vec{y} \otimes \vec{h}) \cdot \matr{m})$. If the weight matrix $\matr{w^{hy}}$ was initialized with zeros, and at the first iteration $\vec{h} = [1, 0, 1, 1]$, $\vec{y} = [1, 0, 1, 0]$, and we set the mask matrix $\matr{m}$ by randomly picking only three among four non-zero connections from the outer product $\vec{y} \otimes \vec{h}$, then the weight matrix $\matr{w^{hy}}$ will be equal to

    $$
        \matr{w^{hy}} = \left( [1, 0, 1, 0] \otimes [1, 0, 1, 1] \right) \cdot \begin{bmatrix}
        1 & 0 & 0 & 0\\
        0 & 0 & 0 & 0\\
        1 & 0 & 1 & 0\\
        0 & 0 & 0 & 0
        \end{bmatrix} = \begin{bmatrix}
        1 & 0 & 0 & 0\\
        0 & 0 & 0 & 0\\
        1 & 0 & 1 & 0\\
        0 & 0 & 0 & 0
        \end{bmatrix}
    $$

 for this particular iteration.

This learning rule, tagged as \textit{simpleHebb}, demonstrates a simple idea of how to bind two binary vectors together but it is impractical because it does not model the forgetting phase (removal of the connections). Next, we will describe models with forgetting while learning from data.

\subsection{Permanence with fixed weight sparsity}
\label{fixed_sparsity}

Removal of the synapses is necessary to avoid the saturation of a binary matrix. This poses a problem: which weights to delete?
If the weights were real-valued, we could prune the weakest synapses.
By the design, however, not due to constraints, our weight matrices are binary. 
Therefore, we introduce a \textit{permanence matrix} $\matr{P}$ that stores real values of how strong the connections are. 
A permanence matrix has the same shape as its binary counterpart and is inspired by the HTM framework \cite{cui2017htm}, where they also operate with permanences.
A permanence matrix is learned continuously from each pair of input-output vectors

\begin{equation}
\label{eq:permanence}
\begin{aligned}
 \matr{P} &= \matr{P} + \lambda (\vec{x_\text{post}} \otimes \vec{x_\text{pre}}) \cdot \matr{m}
\end{aligned}
\end{equation}

and is initialized from a uniform $\mathcal{U}(0, 1)$ distribution. Alternatively, we can initialize $\matr{P}$ with zeros, if the mask matrix $\matr{m}$ has many ones, which means that most of the outer product connections will be updated. $\lambda$ is the learning rate or the plasticity coefficient, which together with $\matr{m}$ defines the learning speed.

Note, although input-output vector pairs update the permanence matrix, the iWTA model eq.\ref{eq:iwta} is still a function of the weight matrix $\matr{w}$, not the permanence.
At a slower time scale, the binary matrix $\matr{w}$ is updated followed by normalization of the permanence matrix $\matr{P}$.
Normalization is an important, otherwise the permanence values grow and new input samples has little effect.
Matrix normalization is performed for each output neuron $i$ individually and is written in the following form:

\begin{equation}
\label{eq:normalization}
\begin{aligned}
  P_{ij} &= \frac{P_{ij}}{\sum_{k} P_{ik}}, \thinspace \forall i,j \\
  \vec{w_i} &= \text{kWTA}(\vec{P_i}, a_w), \thinspace \forall i
\end{aligned}
\end{equation}

Each output neuron forms $a_w$ active synapses that correspond to the largest permanence values (hence the kWTA function is used). This follows Hebbian learning where synapses compete to wire pairs of neurons that are active more often together.
The value of $a_w$ is fixed from the beginning and does not allow the weight sparsity to be changed.
For this reason, we tag this learning rule as \textit{permanence-fixed}.

\subsection{Permanence with varying weight sparsity}
Forcing the sparsity of a permanence matrix to be fixed is a model limitation.
To loose this constraint, we can dynamically change the parameter $a_w$ in eq.\ref{eq:normalization} with learning.
However, if $a_w$ is too high, the encoding sparsity saturates to 1 (see fig.\ref{fig:s_y_s_w}).
 Therefore, we limit the output sparsity of $\vec{h}$ and $\vec{y}$ to be in the fixed range ($s_{min}, s_{max}$) and tune the weights to keep the output sparsity in this desired range.

Let $s_w$ denote the target sparsity of a weight matrix. This is an unknown parameter that is learned with other parameters of a network. We decrease or increase $s_w$ if the encoding sparsity is outside of a given range. A similar mechanism to control the sparsity of encoding is observed in biological neural networks known as homeostatic plasticity \cite{turrigiano2012homeostatic}.

The pseudocode in alg. \ref{alg:varying} adjusts the weight sparsity depending on the output sparsity of excitatory $\vec{y}$ and inhibitory $\vec{h}$ populations. It also modifies the normalization introduced in eq.\ref{eq:normalization} by pruning the permanences that are removed in the weight matrix.
The update of $\matr{P}$ from a sample pair of input-output vectors is the same as in eq.\ref{eq:permanence}. 
The sparsity $s_q$ denotes either $s_y$ or $s_h$.

\begin{algorithm}[h]
\caption{Permanence with varying sparsity}
\label{alg:varying}
\KwIn{the permanence matrix $\matr{P}$, the binary weight matrix $\matr{w}$, the target weight sparsity $s_w$, the matrix row dimension $N_{\text{in}}$ (the number of input neurons), the output sparsity $s_q$, the desired output sparsity range $(s_{\min}, s_{\max})$, the update speed $\gamma$}
\KwOut{$\matr{P}$, $\matr{w}$, $s_w$}
\If{$s_q > s_{\max}$}{
$
s_w \gets \begin{cases}
			\max \left[ 0.05, \thinspace s_w (1 - \gamma) \right], & \text{if \text{excitatory}}\\
            \min \left[ 0.95, \thinspace s_w (1 + \gamma) \right], & \text{otherwise}
		 \end{cases}
$
}
\If{$s_q < s_{\min}$}{
$
s_w \gets \begin{cases}
			\min \left[ 0.95, \thinspace s_w (1 + \gamma) \right], & \text{if \text{excitatory}}\\
            \max \left[ 0.05, \thinspace s_w (1 - \gamma) \right], & \text{otherwise}
		 \end{cases}
$
}
$a_w \gets \left \lceil s_w N_{\text{in}} \right \rceil$ \tcp*{Take the ceil of the product}
$P_{ij} \gets \frac{P_{ij}}{\sum_{k} P_{ik}}, \thinspace \forall i,j$ \tcp*{Normalize $\matr{P}$ by the presynaptic sum}
$\vec{w_i} \gets \text{kWTA}(\vec{P_i}, a_w), \thinspace \forall i$ \tcp*{Leave the $a_w$ largest entries of $\matr{P}$ in $\matr{w}$ for each output neuron}
$\matr{P} \gets \matr{P} \cdot \matr{w}$ \tcp*{Prune permanences, removed in $\matr{w}$}
\end{algorithm}

We limit the value of $s_w$ to be in $(0.05, 0.95)$ range to prevent saturation of a binary weight matrix. At the beginning, $s_w$ is initialized as $s_w \sim \mathcal{U}(0, 1)$ for each matrix. Note, the algorithm deals with excitatory and inhibitory synapses differently.

We will be referring to this learning model as \textit{permanence-varying}. The main distinction from the previous learning model eq.\ref{eq:normalization} is that 1) $a_w$ depends on the output sparsity and 2) the permanences, removed in $\matr{w}$, are also pruned.

The results of applying the proposed learning rules to the iWTA model are presented in the next section.

\section{Results}

\subsection{Metrics}

We define the \textit{convergence} and the \textit{error} as two metrics to evaluate the learning progress.
The convergence is calculated as the average number of elements changed in the output vector compared with the previous iteration:
  \begin{equation}
  \begin{aligned}[c]
    C = \frac{||\vec{y}_i \oplus \vec{y}_{i-1}||_0}{N_y}
  \end{aligned}
  \end{equation}
where $i$ is the iteration and $\oplus$ denotes the XOR operation on binary vectors. The learning process is converged when this value is close to zero.

We define the error function  for a pair of examples $(i, j)$ as:
  \begin{equation}
  \label{eq:error_function}
  \begin{aligned}[c]
    e(i, j) = 
    \begin{cases}
        1 - \cos \left (\vec{y}^{(i)}, \vec{y}^{(j)} \right ) & \text{ if } \text{cluster}^{(i)}=\text{cluster}^{(j)} \\
        \cos \left (\vec{y}^{(i)}, \vec{y}^{(j)} \right ) & \text{ otherwise}
    \end{cases}
  \end{aligned}
  \end{equation}
where $\cos \left (\vec{y}^{(i)}, \vec{y}^{(j)} \right ) $ is the cosine similarity of two vectors. 
This metric is used in the case when the input vector are sampled from different clusters  (see the clustering experiment).  
Then the total error is $E=\sum_{i,j}e(i,j)$. 
During the learning, the cluster index information was not given to the model  - it serves only to evaluate the results.

\subsection{Experimental setup}

The dimensions of the input data $\vec{x}$, output vectors $\vec{h}$ and $\vec{y}$ are set to 200 unless stated otherwise.
We have tested the model with higher dimensions (500 and 1000 neurons), and the results were qualitatively similar.
The input $\vec{x}$ is generated from a Bernouilli distribution with $p=0.2$, although the model can also handle dense $\vec{x}$ vectors with $p=0.5$. 
Initial weight matrices were drawn from a Bernouilli distribution with $p=0.05$, and the permanence matrices were sampled uniformly as shown in section \ref{fixed_sparsity}. 
In the permanence with varying sparsity model (see alg.\ref{alg:varying}), the output sparsity is kept in range $(0.025, 0.1)$, and the update speed $\gamma$ is set to $0.1$. The learning rate is set to 0.01 and, for clustering and decorrelation experiments only, the boolean mask $\matr{m}$ was filled with ones and therefore can be removed from computation in equations \ref{eq:willshaw} and \ref{eq:permanence}.
We will further note that the dependence on the learning rate and the mask $\matr{m}$ may be severe, which can be considered as a model drawback.
An iteration corresponds to a complete run of all data samples (not to be confused with an iteration on fig.\ref{fig:iterations}).

\begin{figure}[t!]
  \begin{subfigure}[t]{0.40\textwidth}
        \includegraphics[width=\linewidth]{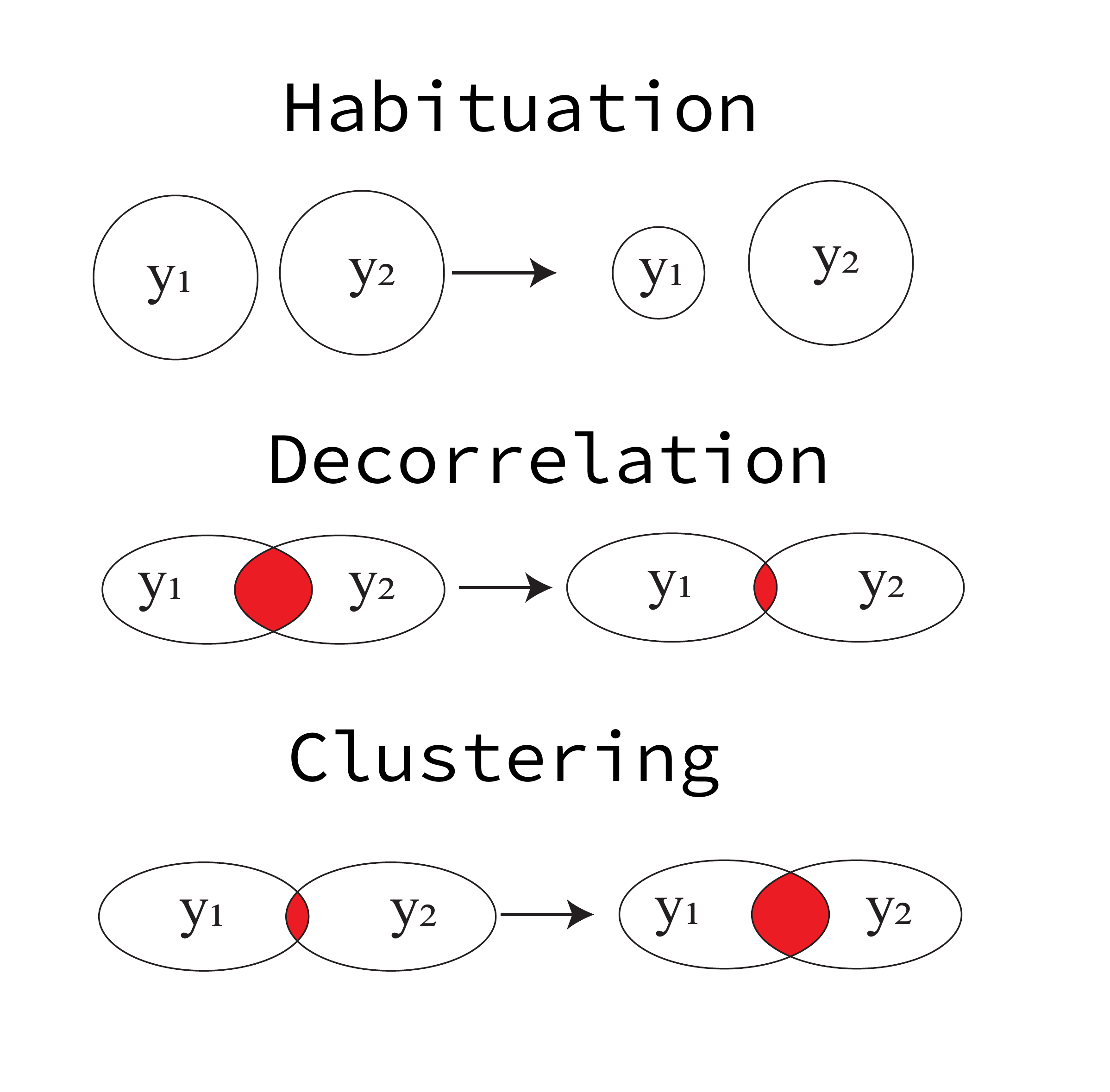}
      \caption{\label{fig:illustration_learning} }  
  \end{subfigure}
   \hfill
  \begin{subfigure}[t]{0.58\textwidth}
        \includegraphics[width=\linewidth]{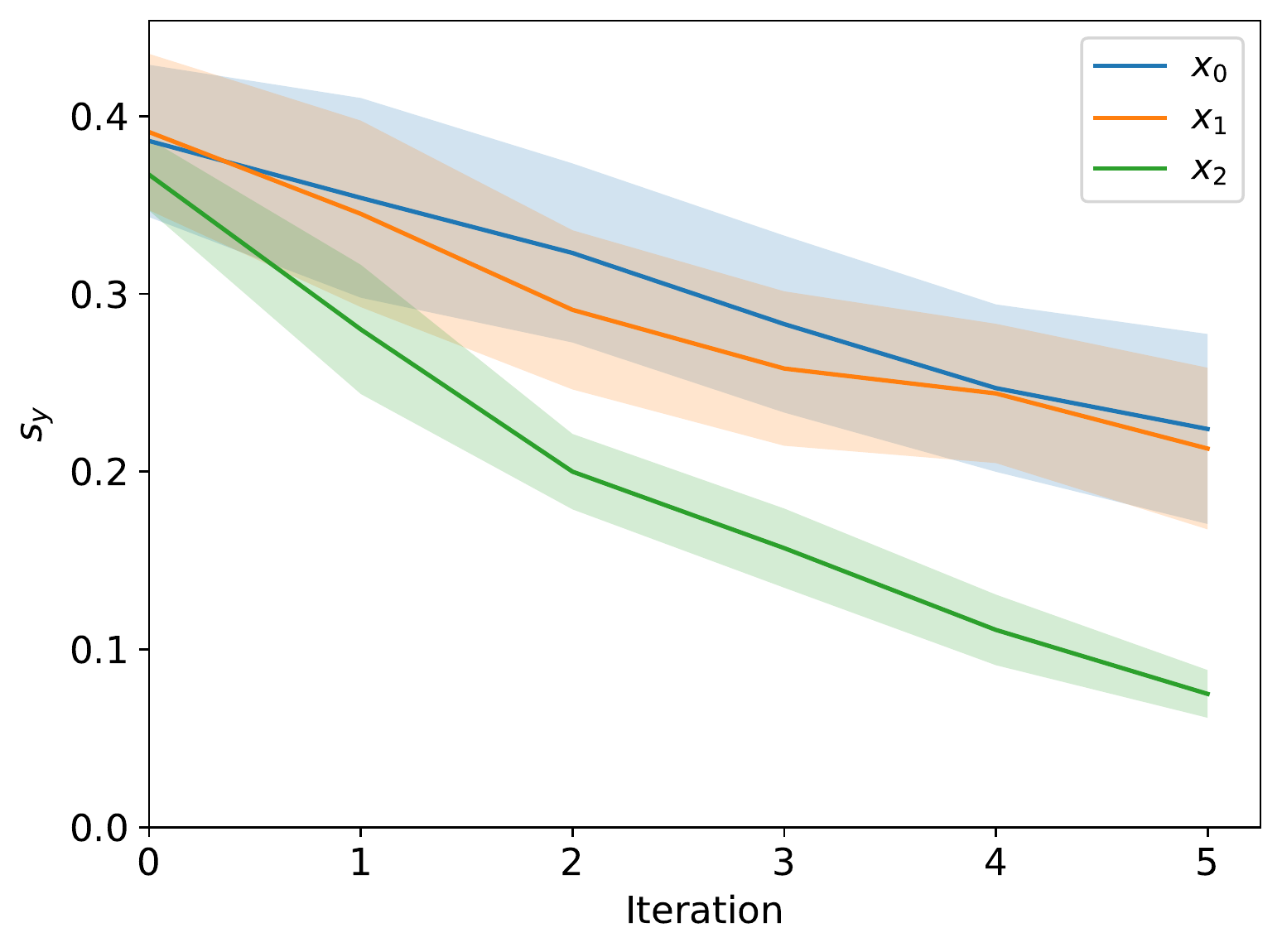}
      \caption{\label{fig:habituation}}
  \end{subfigure}  
         \caption{
         a) The change of the assemblies (in circles) before (left) and after (right) learning. The size of a more frequent population ($\vec{y_1}$) is decreasing, leading to habituation. Decorrelation and clustering decrease and increase the overlap respectively. The opposite effect is achieved by learning other weights.
           b)  	The habituation effect for the \textit{simpleHebb} learning rule without forgetting: the output sparsity $s_y$ encodes information about the frequency of a specific stimulus. 
    Thick lines and filled areas represent the mean and standard deviation accordingly. The input vector $\vec{x_2}$ is encountered more often than $\vec{x_0}$ and $\vec{x_1}$, leading to the sparsity of $\vec{y}(\vec{x_2})$ being smaller than the sparsity of $\vec{y}(\vec{x_0})$  and $\vec{y}(\vec{x_1})$  for the first few iterations. Training $\matr{w^{hy}}$ further will lead to very strong inhibitory signal that suppress any excitation in $\vec{y}$.
         	}
\end{figure}

\subsection{Habituation}

While the kWTA activation function fixes the number of active neurons, the output sparsity of our iWTA model reflects the input distribution: more frequent items will be encoded with less active neurons.
Such adaptation to more frequent stimuli is called habituation.
Fig.\ref{fig:illustration_learning} illustrates the idea of how habituation, decorrelation, and clustering change the population size and the overlap.

In the experiment below, three vectors $\vec{x_0}$, $\vec{x_1}$, and $\vec{x_2}$ are sampled from a Bernouilli distribution with $p=0.2$. 
The frequency with which these vectors are shown in the dataset are 0.2, 0.2 and 0.6 respectively. At each iteration, only 10 synapses from the outer product $\vec{y} \otimes \vec{x}$ in equations \ref{eq:willshaw} or \ref{eq:permanence} are updated (the mask $\matr{m}$ has 10 ones) to show a slow change in the output sparsity.
The learning is applied to $\matr{w^{hy}}$ weights only. Weights $\matr{w^{xh}}$, $\matr{w^{xy}}$, and $\matr{w^{hh}}$ are fixed and initialized from a Bernouilli distribution with $p=0.05$.
Weights $\matr{w^{yy}}$, $\matr{w^{yh}}$, are initialized with lower sparsity ($p=0.01$), since high sparsity of $\matr{w^{yy}}$ rapidly increase the sparsity of the output vector (see fig.\ref{fig:s_y_s_w}),  and high connectivity of $\matr{w^{yh}}$  compensate the effect induced by $\matr{w^{hy}}$ and diminish habituation.

Fig.\ref{fig:habituation} shows how the output sparsity is changing with iteration for the \textit{simpleHebb} learning model. 
After a few iterations, the output sparsity of $\vec{y}(\vec{x_2})$ is less compared to the sparsity of $\vec{y}(\vec{x_0})$ and $\vec{y}(\vec{x_1})$. 
After many iterations, inhibition becomes so strong that the output sparsity for all three input vectors becomes zero.
\textit{Permanence-fixed} learning rule showed similar results.
\textit{Permanence-varying} learning rule did not produce the habituation effect: all output vectors shared similar sparsity.

\subsection{Decorrelation}

In this experiment, we generate 100 random vectors $\vec{x}$ such that each pair of samples $(i,j)$ overlaps in 50\% of active neurons. The recurrent weights $\matr{w^{yy}}$ increase the correlation in $\vec{y}$ between different samples and therefore are removed from the architecture for this experiment. Only $\matr{w^{hy}}$ weights are learned and all other weights are fixed. Under this setup, all learning models presented in section \ref{learning} decrease the overlap between encodings $\vec{y_i}$ and $\vec{y_j}$ equally well for any samples $i$ and $j$ ($i \ne j$). Neurons active in $\vec{y}$ from two or more stimuli are more likely to be inhibited by $\vec{h}$. 
 Thus, the overlap between $\vec{y_i}$ and $\vec{y_j}$ is carved out by inhibition and we observe decorrelation.
 Note, the overlap is decreasing not only because $s_y$ is getting smaller, but also because of the selective $\vec{h} \rightarrow \vec{y}$ inhibitory connections.

\subsection{Clustering}

The clustering experiment aims to show that the iWTA model is able to find clusters in noisy data, such that both $\vec{h}$ and $\vec{y}$ populations are more clustered than noisy input data $\vec{x}$. The experiment can also be viewed as noise removal.

The input data is simulated as follows. 
First, 10 random vectors are sampled from a Bernouilli distribution with $p=0.2$.
We call these vectors \textit{centroids}. 
Then samples around each centroid $\vec{x_c}$ are formed by the XOR operation with binary noise vectors sampled from a Bernouilli distribution with $p=0.1$ (neuron indices are omitted for clarity):
\begin{equation}
\begin{aligned}
    x_c \sim \mathcal{B}(0.2) \\
    n^{(i)} \sim \mathcal{B}(0.1) \\
    x^{(i)} \gets x_c \oplus n^{(i)}
\end{aligned}
\end{equation}
for all $1\le i \le 100$. In total, we have 10 clusters with 100 samples each.

\begin{figure}[t!]
  \begin{subfigure}[t]{0.52\textwidth}
        \includegraphics[width=\linewidth]{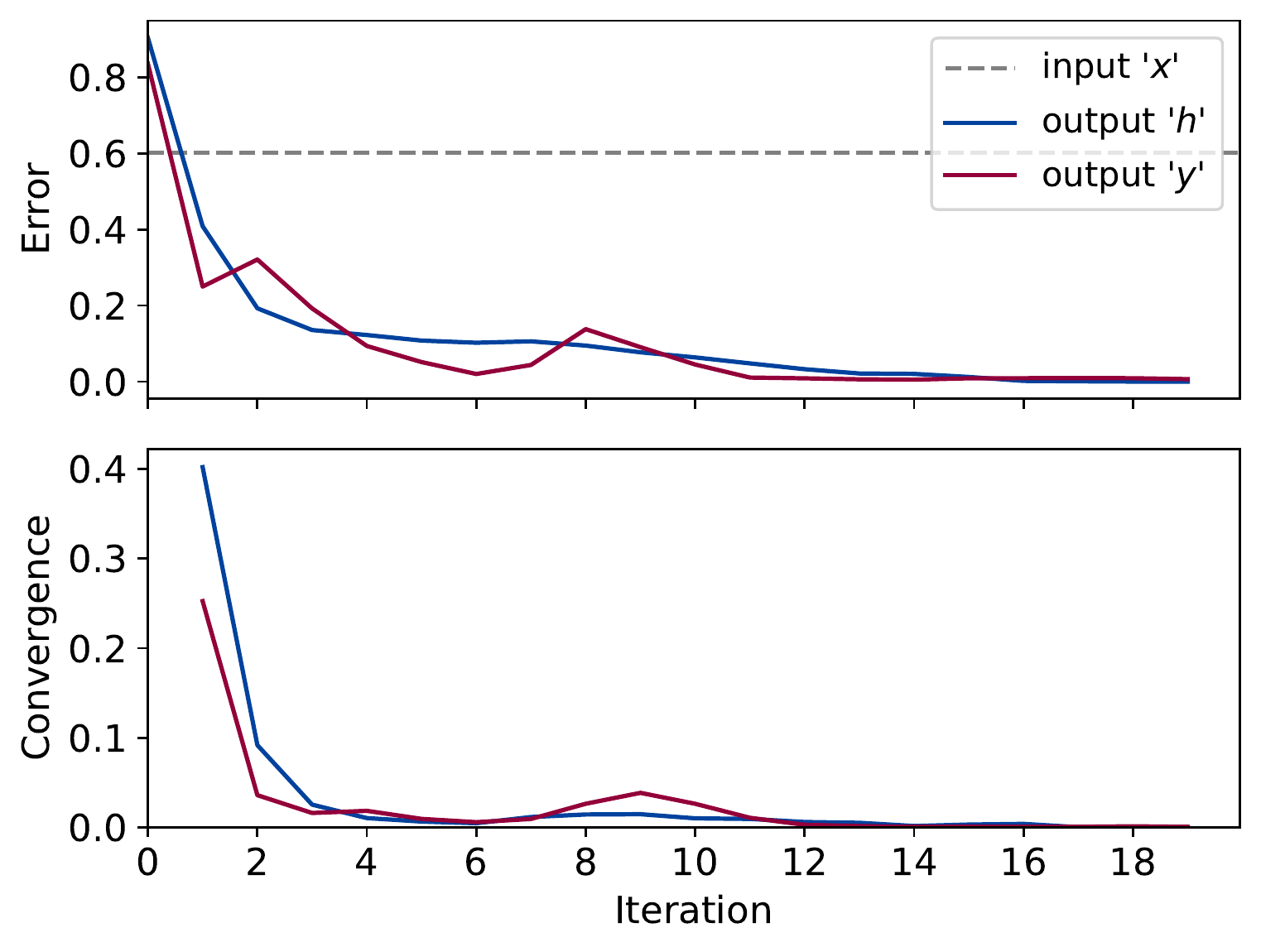}
      \caption{\label{fig:clustering} }
  \end{subfigure}
   \hfill
  \begin{subfigure}[t]{0.48\textwidth}
        \includegraphics[width=\linewidth]{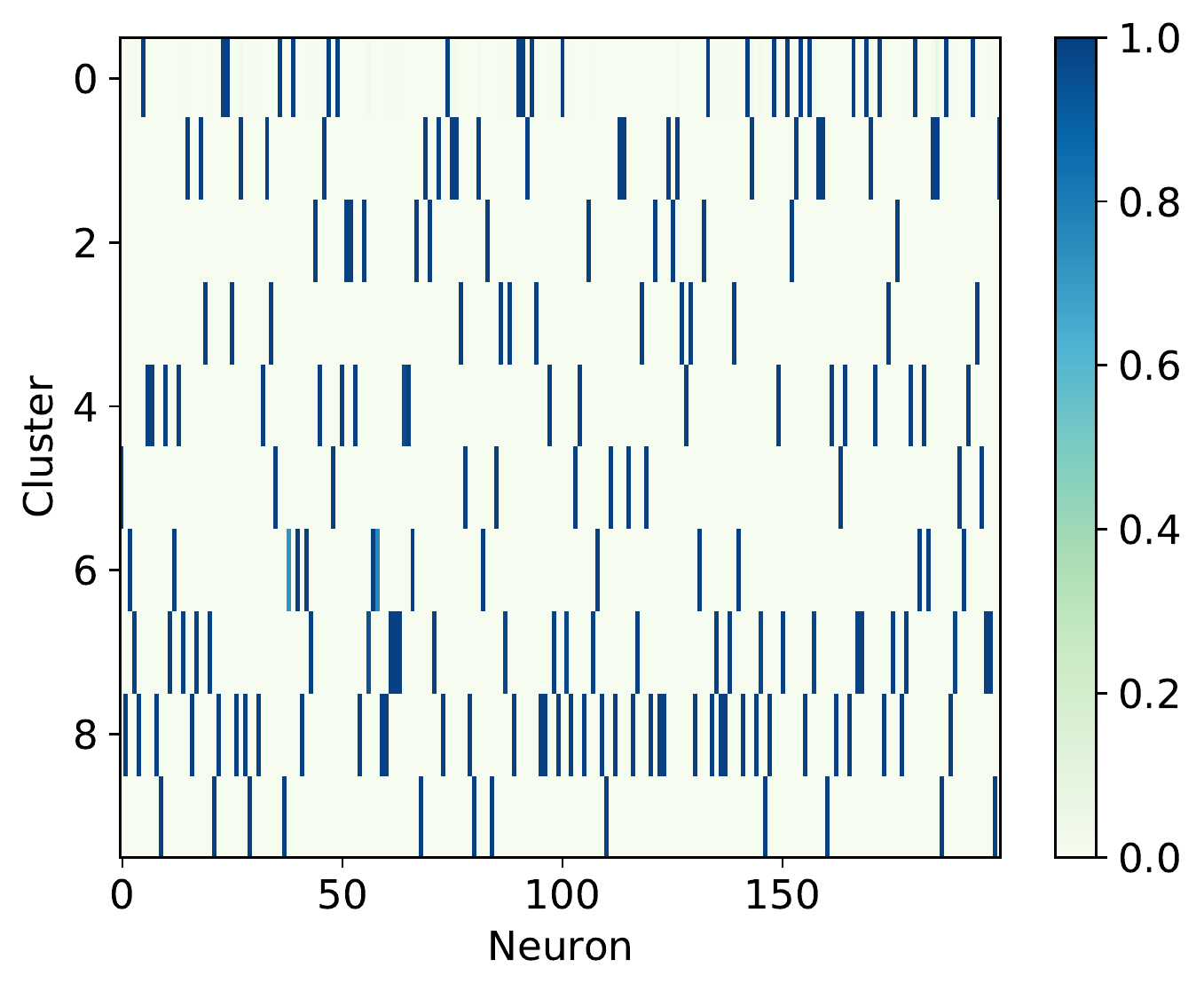}
      \caption{\label{fig:centroids}}
  \end{subfigure}
         \caption{ Clustering noisy data with \textit{permanence-varying} learning rule.
     	a) The model has fully converged after 15 iterations. The error that corresponds to the amount of noise in the data is significantly reduced.
     	b) Activations of $\vec{y}$ averaged across samples for each cluster (vertical axis) after learning. The mean activations for each active neuron inside a cluster approach one (dark blue), meaning that the model has converged. The activations are unique for each cluster.
        }
\end{figure}

In this experiment, all weights - $\matr{w^{xh}}$, $\matr{w^{xy}}$, $\matr{w^{hy}}$, $\matr{w^{hh}}$, and $\matr{w^{hy}}$ - are learned.
As in the preceding experiments with habituation and decorrelation, to learn the weights, we let the model run for a certain number of iterations on $\vec{x}$ to get $\vec{h}$ and $\vec{y}$. At each such iteration, the weights are updated according to the rules in section \ref{learning}.
We found that $\matr{w^{yy}}$ connections force the model output to collapse to a fixed vector independent of $\vec{x}$. Therefore, we have removed $\matr{w^{yy}}$ connections from the iWTA model.
Otherwise, neurons in $\vec{y}$ that are active for two and more $\vec{x}$ from different centroids increase recurrent connectivity $\matr{w^{yy}}$.
This forms a positive feedback loop that quickly degenerates populations $\vec{y}$ into a single vector.

The model with \textit{permanence-varying} learning rule reliably found clusters and is fully converged after a dozen of iterations. The distinct clusters are formed out of noisy realizations of $\vec{x}$.
Fig.\ref{fig:clustering} shows that the noise (the error in eq.\ref{eq:error_function}) is reduced significantly. 
At the same time, different samples from the same cluster are encoded into similar vectors $\vec{y}$. The convergence is observed for all centroids. Fig.\ref{fig:centroids} shows that the active neurons in $\vec{y}$ form clear and distinct clusters from the noisy data. 

Importantly, new distinct input vectors $\vec{x}$ that were not involved in the learning process are encoded differently from the found clusters. Therefore, the model is not limited by the clusters it learned.

Note a small bump in $\vec{y}$ convergence at 9th iteration in fig.\ref{fig:clustering}. That corresponds to a sudden increase in the sparsity of $\vec{y}$.
An increase in the activity of neurons turns on the homeostatic mechanism described in alg.\ref{alg:varying} that reduces the output sparsity by reducing the sparsity of excitatory weights $\matr{w^{xy}}$. This decreases the convergence back to zero as it should be.
If $s_{xy}$ was initially set to a small value, the convergence would not undergo a bump.

The advantage of the model is that it is not sensitive to how the weights are initialized. The model adjusts $s_w$ such that the output sparsity lies in the desired range. Contrary, \textit{simpleHebb} output degenerated into a single vector, and the \textit{permanence-fixed} learning rule failed to find clusters in the input data (the noise was reduced only marginally).

When only $\matr{w^{hy}}$ is learned (as in the previous experiments), the error is not decreased, the noise is not reduced and the model fails to find stable clusters. It is expected behavior, since $\matr{w^{hy}}$ connections are responsible for the decorrelation of the output, reducing the overlap in $\vec{y}$ for all samples from the same cluster.

\subsection{Comparison with the kWTA model}

kWTA does not use explicit inhibition, but we can extend the kWTA model with an additional inhibitory layer from $\vec{h}$ to $\vec{y}$. The populations activity of such a kWTA network can be written as a one-step function:

\begin{equation}
\label{eq:kwta_hy}
\begin{aligned}
\vec{h} &= \text{kWTA}(\matr{w^{xh}}\vec{x}, a_h) \\
\vec{y} &= \text{kWTA}(\matr{w^{xy}}\vec{x} - \matr{w^{hy}}\vec{h}, a_y)
\end{aligned}
\end{equation}

where $a_h$ and $a_y$ are the number of ones in $\vec{h}$ and $\vec{y}$ respectively.

Since the output sparsity of kWTA is fixed, the habituation effect cannot be achieved. For the same reason, the \textit{permanence-varying} learning rule cannot be applied. However, we can compare iWTA with kWTA in two other experiments: decorrelation and clustering.

For the kWTA model, we fixed the output sparsity to be 0.05 for both $\vec{h}$ and $\vec{y}$ populations and learned $\matr{w^{xy}}$ and $\matr{w^{xh}}$ weights only. 
Under the same experimental setup, kWTA achieved the lowest error with the \textit{permanence-fixed} learning rule.
kWTA was able to find clusters in the noisy data, but the best scores kWTA carried out were significantly worse than for the iWTA model. 
In the decorrelation experiment, both models performed equally well.
For a complete comparison of kWTA and iWTA models for clustering and decorrelation, refer to supplementary figures at \cite{Github2021}.

\section{Discussion}
The presented iterative winners-take-all model introduces explicit inhibitory neurons that act as an input-dependent threshold. This allows to better shape cell assemblies with learning. kWTA uses the same threshold for all neurons and iWTA allows this threshold to be different. But it is not the same as an individual threshold (or bias) as in standard artificial neural networks. Classical approaches use the threshold as a parameter that is learned in the training phase and fixed in the test. But iWTA allows this threshold to vary via inhibitory neurons like in biological networks. Furthermore, this threshold depends not only on bottom-up input but also on possible contextual or top-down signals. 

iWTA is compatible with other models that approximate brain computation through connected areas represented as binary vectors like in \cite{papadimitriou2020brain} and \cite{hawkins2019framework}. However, with iWTA, different areas can selectively influence the output encodings by gating, modifying, or predicting the response, which is not possible to do with a simple kWTA model. Also, with iWTA, it is possible to implement ideas from predictive coding, where higher hierarchical areas inhibit the predicted response from lower areas \cite{Spratling2017ARO}. Biologically, it leads to saving energy for signal transmission. Computationally, it also can be used to improve the mutual information between areas. 

In the model description, we constrained activations and weights to be binary. The motivation was that it better corresponds to the cell assembly idea, and learning is interpreted as deletion and formation of connections. However, iWTA also works if the input and weight matrices take real values. Thus, it can be used with arbitrary input to encode it into a binary vector.
Although not tested, we hypothesize that applying learning rules for real-valued weights would lead to similar results that we achieved. 
Increasing the magnitude of real weights $\matr{w^{hy}}$ would similarly increase inhibition and lead to more sparsely active excitatory populations and potentially recreate habitation and decorrelation.

In this work, we presented two mechanisms with permanences that perform local unsupervised Hebbian learning. The benefit of the \textit{permanence-varying} rule is that it is not sensitive to how the weights are initialized (particularly, their initial sparsity). It also converges faster than \textit{permanence-fixed}, which may or may not converge at all. Dense input vectors $\vec{x}$ are also better handled with \textit{permanence-varying}; the habituation effect, however, was not observed with it.

We note that the choice of the learning rate $\lambda$ and how much the boolean mask $\matr{m}$ is filled with ones influences the results.
A large learning rate makes the model unable to stabilize. Small values of the learning rate stop the training preliminary due to quick convergence but may result in higher error. The authors in \cite{hsieh2018optimizing} suggest varying the learning rate through time. A widely used technique is to start with a large learning rate and decrease it with each iteration. This gives a boost to the model not to settle on randomly initialized matrices at the beginning and ensures smooth convergence at the end. 
A wrong learning rate and $\matr{m}$ also increase the chance to degenerate the output into a single vector.
To alleviate this erroneous behavior, the authors in \cite{cui2017htm} keep track of the recent activity of the neurons via a boost factor that is multiplied by the presynaptic sum to allow other neurons to vote. 
It increases the entropy of neural responses and thus the ability to preserve information about the stimuli.

The iWTA model is on a higher abstraction level than a spiking network. Most prominently, it does not use continuous time. Still, active neurons at different iterations can be considered as neurons active at different times. Therefore, we made experiments with spike-timing-dependent plasticity rules. Particularly, we were interested in learning synapses of inhibitory neurons. We followed the idea from  \cite{vogels2011inhibitory} to potentiate inhibitory connections between two neurons if they fire close in time and depress if the spike interval is large. 
Treating the spike interval as the distance between iterations, iWTA showed similar results compared to the \textit{permanence-fixed} learning model and yielded no further improvements.

Overall, iWTA is an interesting and computationally rich model of cell assembly formation. Still, we report a problem with learning rules. Although the presented learning mechanisms are suited for demonstration of habituation, decorrelation, and clustering, they are sensitive to learning rates, experimental setup, and the choice of the weight matrices to learn. Probably, a more solid approach is to learn weights by maximizing the mutual information between layers \cite{Ganguli2014EfficientSE} or minimizing the prediction error \cite{Teufel2020FormsOP}.

\section{Conclusions}
The presented iterative winners-take-all model encodes an arbitrarily input vector into a binary vector that represents a cell assembly. 
The main features of the model are:
inhibition is modeled explicitly compared to kWTA;
excitation-inhibition balance is formed through an iterative procedure;
the input similarity is preserved as in kWTA;
paradoxical effects are visible in excitation-inhibition interaction.

Simple Hebbian-like learning rule without forgetting demonstrates a simple idea of how to learn synapses between two populations but has problems with learning binary weights from data.
The habituation effect emerges by learning the inhibitory connections $\matr{w}^{hy}$ in the iWTA model, which is not possible in kWTA models.
By introducing a permanence matrix, we were able to prune the weights and achieved clustering and decorrelation effects.
Still, the problem of learning binary connections is not solved and requires future work.

\bibliographystyle{unsrt} 
\bibliography{reference}

\end{document}